\pdfoutput=1

\documentclass{article}
\usepackage{spconf,amsmath,graphicx}
\usepackage{color}
\usepackage{enumitem}
 \usepackage{booktabs}
 \usepackage{graphicx}
 \usepackage{amssymb}
\usepackage{hyperref}
\hypersetup{
    colorlinks=true,
    linkcolor=blue,
    filecolor=magenta,      
    urlcolor=cyan,
}

\setlist{nosep, leftmargin=14pt}

\usepackage{mwe} 

\title{Exploring Instance-Level Uncertainty \\ for Medical Detection}

 \name{Jiawei Yang$^{\star,1}$\thanks{Jiawei Yang and Yuan Liang contributed equally and were considered as co-first authors($\star$).}, Yuan Liang$^{\star,1}$, Yao Zhang$^{2,3}$, Weinan Song$^1$, Kun Wang$^{1}$, Lei He$^{1}$}
 \address{$^{1}$ Electrical and Computer Engineering, University of California, Los Angeles, CA, USA \\ 
 $^{2}$Institute of Computing Technology, Chinese Academy of Sciences, Beijing, China\\
 $^{3}$University of Chinese Academy of Sciences, Beijing, China}

\begin{document}
%
\maketitle
\vspace{-0.5cm}
\begin{abstract}
The ability of deep learning to predict with uncertainty is recognized as key for its adoption in clinical routines. 
Moreover, performance gain has been enabled by modeling uncertainty according to empirical evidences.   
While previous work has widely discussed the uncertainty estimation in segmentation and classification tasks, its application on bounding-box-based detection has been limited, mainly due to the challenge of bounding box aligning. 
In this work, we explore to augment a detection CNN with two different bounding-box-level (or instance-level) uncertainty estimates, \textit{i.e.}, predictive variance and Monte Carlo (MC) sample variance. 
Experiments are conducted for lung nodule detection on LUNA16 dataset, a task where significant semantic ambiguities can exist between nodules and non-nodules. 
Results show that our method improves the evaluating score from 84.57\% to 88.86\% by utilizing a combination of both types of variances. 
Moreover, we show the generated uncertainty enables superior operating points compared to using the probability threshold only, and can further boost the performance to 89.52\%. 
Example nodule detections are visualized to further illustrate the advantages of our method.
Our implementation can be found in \url{https://git.io/JTwhe}. 
\end{abstract}

%
%
\vspace{-0.2cm}
\section{Introduction}
\label{sec:intro}
Medical image diagnosis with convolutional neural networks (CNNs)  is still challenging due to the semantic ambiguity of pathologies \cite{liang2020atlas}. 
For example, lung nodules and non-nodules can share instinctive similarity, which can lead to high false-positive rates from 51\% to 83.2\% according to experts' inspection \cite{harsono2020lung}. 
As such, the ability of CNN to quantify uncertainty has recently been identified as key for its application in clinical routine to assist clinicians' decision making and gain trusts 
\cite{amodei2016concrete}. 
Recently, approximating the posterior of a CNN using dropout and MC samples provides a simple approach to estimate uncertainty \cite{gal2016dropout}, and has been applied for several medical image diagnosis tasks. 
As for classification tasks, example works include the modeling of uncertainty for diabetic retinopathy diagnosis from fundus images \cite{leibig2017leveraging} and lesion detection from knee MRI \cite{pedoia20193d}.  
As for segmentation tasks, uncertainty estimation has been applied to localize lung nodules \cite{ozdemir2017propagating}, brain sclerosis lesion \cite{nair2020exploring}, brain tumor \cite{wang2019aleatoric}, \textit{etc.}. 
Moreover, performance boost has been observed from the above studies by utilizing uncertainty as a filter for false positive predictions besides the probability thresholding. 

Despite the aforementioned work, only limited studies have explored the uncertainty estimation for detection task. 
The main challenge is that, while MC samples from segmentation and classification are naturally well aligned, bounding box samples from the detection task are spatially diverse, and must be associated before aggregation \cite{miller2019evaluating}. 
One existing solution is to derive pixel-level uncertainty from segmentation, and then aggregate pixels of a connected region with a log-sum operation for instance-level uncertainty \cite{nair2020exploring}.    
However, the method cannot optimize uncertainty in an end-to-end manner, and the log-sum aggregation has the potential stability issue. 
Another solution is to define bounding box clustering strategies for merging box samples of a same object, \textit{e.g.} affinity-clustering \cite{miller2019evaluating} and enhanced Non-Maximum Suppression (NMS) \cite{he2019bounding} as proposed for estimating boundary uncertainty from natural photos. 
However, the methods require extra clustering parameters from handcrafting, and the optimal values can depend on specific tasks. 

\begin{figure*}[ht!]
\begin{center}
\includegraphics[width=0.75\linewidth]{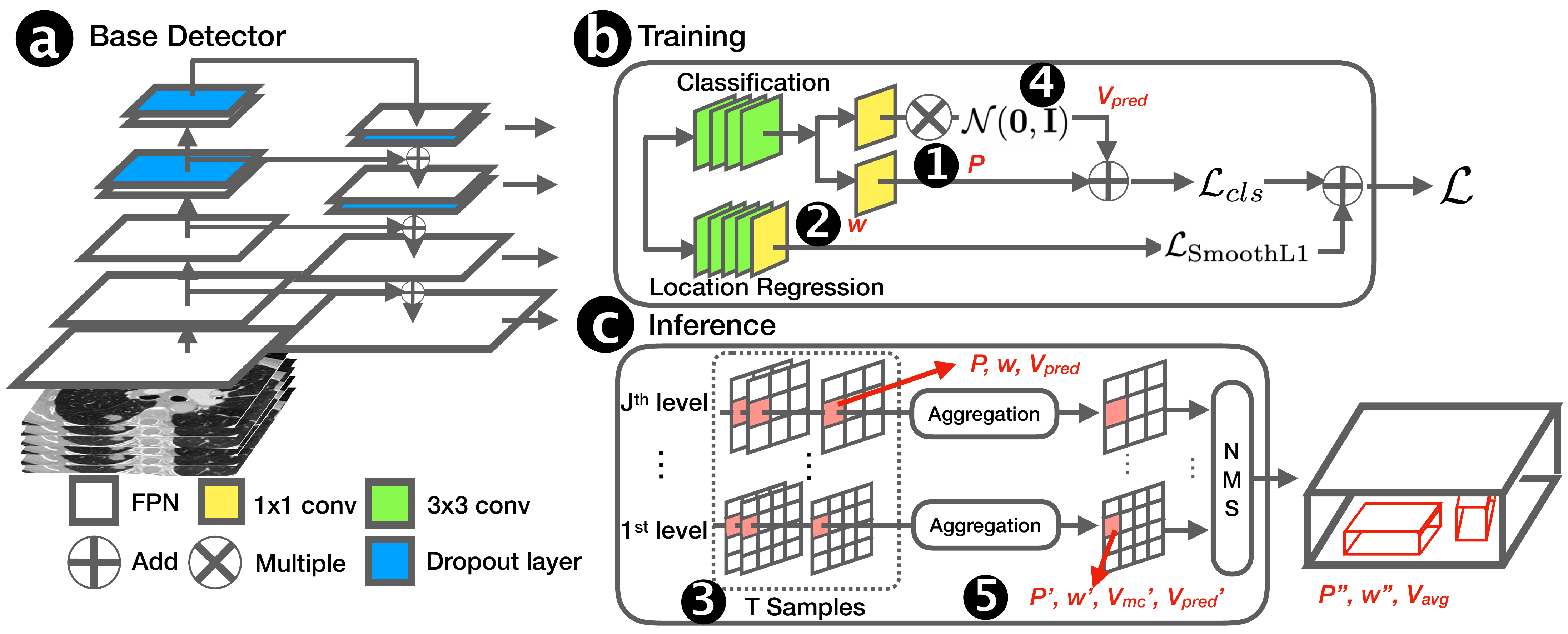}
\vspace*{-\baselineskip}
\end{center}
   \caption{Model overview. (a) a single-scale multi-level pyramid CNN with dropout is used as the detector. (b) During training, bounding box predictions of probability, predictive variance, and location parameters are trained directly against ground-truth. (c) During inference, MC samples of bounding box for each pyramid level are first in-place aggregated for MC variances, which further averaged with predictive variances as the uncertainty estimation.}
\label{fig:overview}
\vspace*{-\baselineskip}
\end{figure*}

Different from previous work, we propose to estimate instance-level uncertainty directly from a single-scale multi-level detector in an end-to-end manner.
Specifically, since object occlusion is rare in medical analysis, a single-scale bounding box prediction is generated at each pyramid level, which enables the simple clustering of multiple MC samples without the need for alignment.
Two types of uncertainty measure, \textit{i.e.} predictive variance and MC sample variance, are studied. 
Our experiments show that: (\textit{i}) a combination of both types of uncertainty leads to best performance, and (\textit{ii}) using uncertainty as a bounding box inclusion criteria besides probability allows superior operating points.

%

\vspace{-3mm}%

\section{methods}
\vspace{-0.2cm}
Figure \ref{fig:overview} shows the overview of our method. 
We develop a 2.5D single-scale multi-level pyramid CNN with dropout to predict bounding boxes of pathologies with attributes of probability and uncertainty from an input image.
During training, box-wise probability, predictive variance and location parameters are directly trained with ground-truth supervision. 
During testing, an unseen image is passed through the model $T$ times to estimate MC variance. 
With the single-scale structure, both MC and predictive variances of bounding boxes can be simply averaged in-place for aggregation. 


\vspace{-0.2cm}
\subsection{Single-Scale Multi-Level Detector}
\label{sec:det}

A Feature Pyramid Network (FPN) similar to \cite{lin2017focal} is applied to extract multi-scale features as shown in Figure \ref{fig:overview}(a). 
By referring to existing medical detectors \cite{zhang2018sequentialsegnet,khosravan2018s4nd,liang2020oralcam}, the input volumes are designed to be stacked 2D slices of a 3D image for model efficiency, referred as 2.5D. 
By following the  \cite{liu2016ssd}, bounding box predictions are generated from the multi-level feature maps for label probability $P$ (Figure \ref{fig:overview}(1)) and location/size deltas $\omega$ (Figure \ref{fig:overview}(2)). 
Different from detectors for natural scenarios, only one base scale (aspect ratio and size) for bounding boxes is defined at the each level by considering that the pathologies in medical domain can be more regular in shapes. 
The model can also be viewed as a multi-level extension of S4ND \cite{khosravan2018s4nd}, a CNN detector with a single-scale bounding box design for nodule detection.
The base bounding box sizes at different levels can be designed to be sufficient to fit all the expected pathology sizes in a target task.  



\vspace{-0.2cm}
\subsection{Monte Carlo Variance}
\vspace{-0.1cm}
Previous studies show that minimizing the cross-entropy loss of a network with dropout applied after certain layers enables capturing the posterior distribution of the network weights \cite{gal2016dropout}. 
We follow the method by enabling dropout operations during both training and testing stages. 
For each inference, $T$ forward passes of a target volume are conducted, which results in $T$ sets of bounding box MC samples with $P$ and $\omega$ at each pyramid level (Figure \ref{fig:overview}(3)). 
As such, MC variance $V_{mc}$ can be defined as the variance of the $P$ for all associated MC samples of a bounding box, and is used as a measure of an instance-level uncertainty. 

\vspace{-0.3cm}
\subsection{Predictive Variance}
\vspace{-0.1cm}

During training, the weights of the network are also trained to directly estimate a bounding-box-wise variance $V_{pred}$ (Figure \ref{fig:overview}(4)) by following the method of \cite{kendall2017uncertainties}. 
In specific, by assuming the classification logits $\mathbf{z}$ are corrupted by a Gaussian noise with variance $\boldsymbol{\sigma}$ at each bounding box prediction, the weight updates during back-propagation encourages the network to learn the variance estimates without having explicit labels for them. Defining the corrupted output $\mathbf{z}_{i,t} = \mathbf{z}_i + \boldsymbol{\sigma}_i\times\boldsymbol{\epsilon}_t,\boldsymbol{\epsilon}_t \sim \mathcal{N}(\mathbf{0},\mathbf{I})$, the loss function can be written as $\mathcal{L}_{cls} = \sum_{i}\log\frac{1}{T'}\sum_{t} \mathrm{softmax}(\mathbf{z}_{i,t})$,
where the $T'$ is the number of times for MC integration. 
Since the method models object ambiguity during the optimization, it has shown to enable improved performance \cite{nair2020exploring,kendall2017uncertainties}. 
During the inference, each MC sample of a bounding box prediction comes with a predictive variance $V_{pred}$, and is used as another uncertainty estimation.  



\begin{figure*}[ht!]
\begin{center}
\vspace*{-\baselineskip}
\includegraphics[width=\linewidth,height=7.5cm]{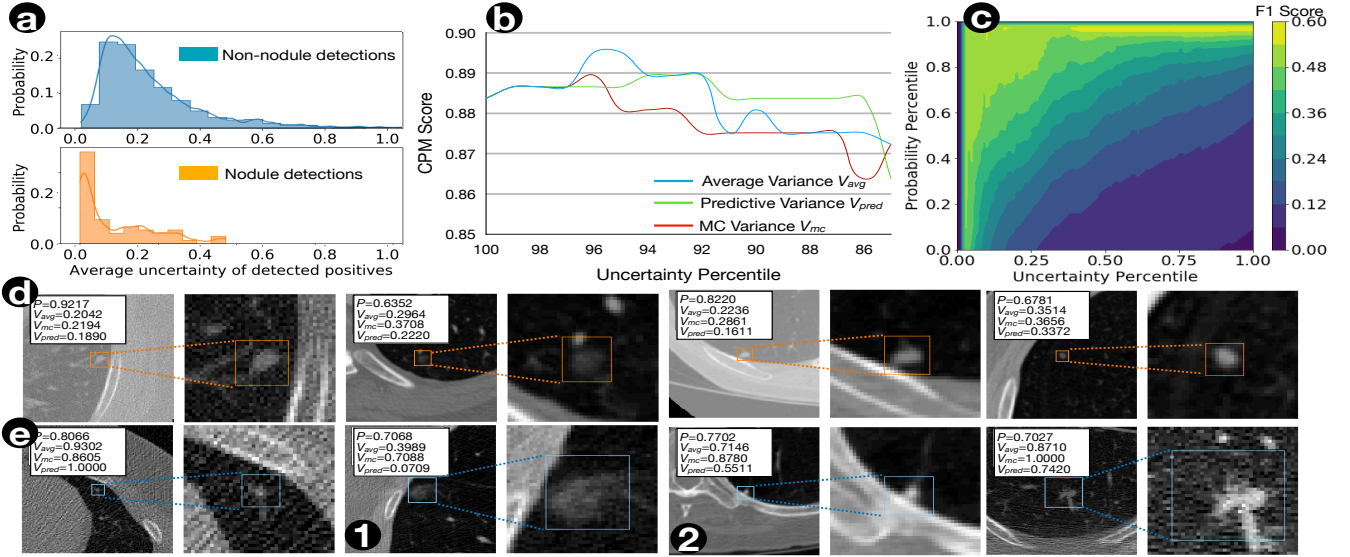}
\vspace*{-\baselineskip}
\end{center}
  \caption{(a) Estimated uncertainty ($V_{avg}$) distribution of nodule and non-nodule detections. (b) CPM curves under different uncertainty thresholds. (c) F1 score under different probability and uncertainty thresholds. (d) Examples of nodule detections. (e) Examples of non-nodule detections. }
\label{fig:threshold}
\vspace*{-\baselineskip}
\end{figure*}

\vspace{-0.3cm}%
\subsection{MC Sample Aggregation} 
All the bounding boxes samples from multiple MC inference in each pyramid level are aggregated in-place as shown in Figure \ref{fig:overview}(5). 
In specific, the probability $P$ and location parameters $\omega$ of a bounding box at any grid point can be obtained by averaging those of all the MC samples that are at the same location.
The MC variance of a bounding box can be calculated as $V_{mc} = (\sum_{t}\mathbf{P}_t^2)/T - (\sum_{t}\mathbf{P}_t/T)^2$, where $P_t$ represents the predicted probability of a MC sample at the location. 
The predictive variance of a bounding box can be represented as $V_{pred} = \sum_t\mathrm{softmax}(\boldsymbol{\sigma}_t^2)/T$, where $\boldsymbol{\sigma}_t$ is the predictive variance output for each MC sample. 
The aggregated bounding boxes among different pyramid levels are then post-processed with regular NMS for removing overlaps. 
The final variance for a predicted bounding box is taken as the average of its predictive variance and MC variance, represented as $V_{avg} = (V_{mc} + V_{pred}) / 2$.

\vspace{-0.3cm}
\section{Experiments}
\vspace{-0.2cm}
\subsection{Dataset and Metrics}
We evaluated the proposed method for the lung nodule detection task on LUNA16 dataset~\cite{setio2017validation}, 
where 200/30/30 scans were randomly selected for training, validation, and testing. 
As suggested by the original challenge, we employed the Competition Performance Metric (CPM) to evaluate the model performance, which is defined as the mean of sensitivities at the average false positives (FPs) of 0.125, 0.25, 0.5, 1, 2, 4, and 8 per scan. 
Moreover, we also utilized F1-score as the nodule detection performance to study the effect of using different probability and uncertainty as thresholds.

\vspace{-0.3cm}
\subsection{Implementation}
\vspace{-0.2cm}

We set the size of an input patch to be 228$\times$228$\times$7. 
The base sizes of bounding box predictions are set to be 8$\times$8$\times$7, 16$\times$16$\times$7, 32$\times$32$\times$7, and 64$\times$64$\times$7 at the four pyramid levels in order to fit most nodules.  
We set a dropout rate of 0.1 for all dropout layers, and the number MC sampling $T$ to be 10 for inference.
All volumes from LUNA16 were re-sampled to a voxel spacing of 0.7mm$\times$0.7mm$\times$1.25mm, are are clipped to intensity range of $[-1200, 800]$ as common practice in lung nodule detection problem\footnote{LUNA16 winner: \url{shorturl.at/coqI7} and another work\cite{jaeger2020retina} for LIDC.}.
Intensive augmentations of scaling, rotation, shifting, and random noise were applied during training. All models are trained from scratch for 200 epochs using Adam optimizer at the initial learning rate of $10^{-4}$. The baseline network design closely follows \cite{jaeger2020retina}.
\vspace*{-\baselineskip}
\begin{table}[ht!] 

\centering 
\caption{Performance comparison between different models. M1-3 are our networks with different variance potentials, while Unet1-3 are segmentation-based detection models.}
\label{tab:main_results}
\resizebox{\columnwidth}{!}{%
\begin{tabular}{@{}c|c|c|c|c@{}}
\toprule
Methods			& V$_{mc}$ & V$_{pred}$   & Loss Function & CPM($\%$)        \\ \midrule\midrule
Liao et al.~\cite{liao2019evaluate}   & $\times$& $\times$ & --    & 83.4    \\
Zhu et al.~\cite{zhu2018deeplung}    & $\times$& $\times$ & --    & 84.2    \\
Li et al.~\cite{li2019deepseed:}     &$\times$ & $\times$ & --    & 86.2    
\\\midrule
Unet1   & $\times$ & $\times$        & $\mathcal{L}_{\mathrm{CE}}+\mathcal{L}_{\mathrm{Dice}}$   & 78.78 --         \\
Unet2  & \checkmark & $\times$        & $\mathcal{L}_{\mathrm{CE}}+\mathcal{L}_{\mathrm{Dice}}$   & 80.00 $\uparrow$         \\
Unet3    & \checkmark & \checkmark        & $\mathcal{L}_{\mathrm{cls}}+\mathcal{L}_{\mathrm{Dice}}$   & 80.13 $\boldsymbol{\uparrow}$         \\\midrule

M1   & $\times$ & $\times$        & $\mathcal{L}_{\mathrm{CE}}+\mathcal{L}_{\mathrm{SmoothL1}}$   & 84.57 --         \\
M2   & \checkmark & $\times$        & $\mathcal{L}_{\mathrm{CE}}+\mathcal{L}_{\mathrm{SmoothL1}}$   & 87.14 $\uparrow$          \\
M3 & \textbf{\checkmark} & \checkmark & $\mathcal{L}_{cls}+\mathcal{L}_{\mathrm{SmoothL1}}$ & 88.86 $\boldsymbol{\uparrow}$ \\
M3$_{\eta=0.456}$ & \checkmark & \checkmark & $\mathcal{L}_{cls}+\mathcal{L}_{\mathrm{SmoothL1}}$ & \textbf{89.52} $\boldsymbol{\uparrow}$
\\ \bottomrule
\end{tabular}%
}
\vspace*{-\baselineskip}
\end{table} 

%


\vspace{-0.3cm}
\subsection{Results and Discussion}
\vspace{-0.1cm}%
\subsubsection{Performance Comparison}
\vspace{-0.1cm}
Table~\ref{tab:main_results} shows the performance of different models.
Multiple recent detection-based networks without estimating uncertainty \cite{liao2019evaluate,setio2016pulmonary,zhu2018deeplung,li2019deepseed:} is used as the baseline models for comparison. 
Unet1-3 are segmentation-based models that draw detection output by merging connected pixels by following the strategy of \cite{nair2020exploring}: Unet1 exploits the nnUNet \cite{isensee2019nnu-net:} without modeling uncertainty, Unet2 extends Unet1 with MC variance, and Unet3 models both MC and predictive variance. 
M1-3 are our detection models for enabling bounding-box-level uncertainty: M1 is the base detector without uncertainty measure, while M2 has MC variance and M3 is the full model that enables both MC and predictive variance. 
We can see that our models (M1-3) achieve comparable results with existing detection-based methods. 
In specific, MC inference improves CPM by 2.57\% with dropout layers incorporated (M1$\rightarrow$M2), and $V_{pred}$ further improves CPM by 1.72\% by introducing predictive variance in the optimization process. 
Moreover, we empirically set the uncertainty threshold of 0.456 (96\% percentile) from search in validation set to filter out bounding box predictions with a high uncertainty, where the optimal threshold value is determined based on the validation set. The extra filter further boosts the model performance by 0.66\% in CPM (M3$\rightarrow$M3$_{\eta=0.456}$).
By comparing our method with uncertainty estimation from merging segmentation results (Unet1-3), we can see that deriving bounding-box-wise uncertainty in an end-to-end manner enables higher CPM boosts. 

\vspace{-0.3cm}
\subsubsection{Exploiting Uncertainty Information and Case Study}
Figure \ref{fig:threshold}(a) shows the distribution of the estimated uncertainty (using a combination of $V_{pred}$ and $V_{mc}$) of positive detections and negative detections. 
We can see that bounding boxes for non-nodules have a generally higher uncertainty level than nodules, which indicates that the uncertainty estimation can be used to improve model performance by filtering out false positive findings. 
In Figure \ref{fig:threshold}(b), we study the model performance under different uncertainty thresholds on the testing set: the CPM score of the model first increases mainly due to the filtering out of false positive detections; and then it decreases possibly because certain true positive findings are mis-removed. 
We also observe that using a combination of $V_{pred}$ and $V_{mc}$ leads to the highest CPM, which indicates the two types of variance can complement for the improved performance. 
We further plot the model performance as F1 score under different values of uncertainty and probability thresholds in Figure\ref{fig:threshold}(c), which also confirms that that using both parameters as thresholds lead to the optimal performance. 
In the real application, the uncertainty and probability thresholds can be tuned together to meet the specific precision and recall requirements of a task. 
Figure \ref{fig:threshold}(d,e) visualizes example bounding-box detections with the estimated $p$, $V_{avg}$, $V_{pred}$, and $V_{mc}$. 
In specific, Figure \ref{fig:threshold}(d) includes true positive detections with outputs of high probability and low averaged uncertainty, where the model is confident about the prediction; Meanwhile, Figure \ref{fig:threshold}(e) includes non-nodule detections with high estimated probability but high uncertainty, and are correctly filtered out by our model by using uncertainty as a threshold.
Moreover, cases show that predictive and MC variances can capture different uncertainties and complement each other (Figure \ref{fig:threshold}(1,2)), which also validates our method of combining both types of variance.   

\vspace{-0.4cm}%
\section{Conclusion}
\vspace{-0.2cm}
In this work, we propose to estimate instance-level uncertainty in an end-to-end manner with a single-scale multi-level detection network.
Two types of uncertainty measures, \textit{i.e.}, predictive variance and MC variance, are studied.
Experimental results prove that the incorporation of uncertainty estimation improves model performance, can act as filters for false positive detections, and can be used with probability as thresholds for setting superior operating points.  

\vspace{-0.2cm}
\section{Acknowledgments}
\vspace{-0.2cm}

This manuscript has no conflict of interest.

\vspace{-0.2cm}
\section{Compliance with Ethical Standards}
\vspace{-0.3cm}

This research study was conducted retrospectively using human subject data made available in open access by LUNA16\footnote{https://luna16.grand-challenge.org/}. The related paper is \cite{setio2017validation}. Ethical approval was not required as confirmed by the license attached with the open access data.

\vspace{-0.3cm}

\bibliographystyle{IEEEbib}
\bibliography{strings,refs}

\end{document}